\documentclass{article}

\PassOptionsToPackage{numbers, compress}{natbib}

\usepackage[preprint]{neurips_2024}

\usepackage[utf8]{inputenc} % allow utf-8 input
\usepackage[T1]{fontenc}    % use 8-bit T1 fonts
\usepackage{hyperref}       % hyperlinks
\usepackage{url}            % simple URL typesetting
\usepackage{booktabs}       % professional-quality tables
\usepackage{amsfonts}       % blackboard math symbols
\usepackage{nicefrac}       % compact symbols for 1/2, etc.
\usepackage{microtype}      % microtypography
\usepackage{xcolor}         % colors
\usepackage{tablefootnote}
\usepackage{amsmath}
\usepackage{graphicx}

\usepackage{multirow}
\title{Transparent Object Depth Completion}

\author{%
  Yifan Zhou, Wanli Peng, Zhongyu Yang, He Liu, Yi Sun\thanks{Corresponding author.} \\
  Dalian University of Technology \\
  \{xyzyf, 1136558142, yzyyang, 11909016\}@mail.dlut.edu.cn, {lslwf@dlut.edu.cn} \\
}

\begin{document}

\maketitle

\begin{abstract}
The perception of transparent objects for grasp and manipulation remains a major challenge, because existing robotic grasp methods which heavily rely on depth maps are not suitable for transparent objects due to their unique visual properties. These properties lead to gaps and inaccuracies in the depth maps of the transparent objects captured by depth sensors. To address this issue, we propose an end-to-end network for transparent object depth completion that combines the strengths of single-view RGB-D based depth completion and multi-view depth estimation. Moreover, we introduce a depth refinement module based on confidence estimation to fuse predicted depth maps from single-view and multi-view modules, which further refines the restored depth map. The extensive experiments on the ClearPose and TransCG datasets demonstrate that our method achieves superior accuracy and robustness in complex scenarios with significant occlusion compared to the state-of-the-art methods.
\end{abstract}

\section{Introduction}
Currently significant achievements have been made in the field of robotic grasp and manipulation \cite{27chen2023efficient,28barad2023graspldm,29tang2023graspgpt}. However, the perception of transparent objects for grasp and manipulation remains a major challenge in this field. Transparent objects, such as glass cups and plastic bottles, are common in daily life, yet most existing methods perform poorly on them. These methods \cite{27chen2023efficient,28barad2023graspldm,29tang2023graspgpt} are highly dependent on the depth maps produced by depth sensors, but optical depth sensors struggle to accurately capture the depth of transparent objects. The unique visual properties of transparent objects result in specific challenges for optical sensors, where specular reflection at their surfaces leads to gaps in depth maps and light refraction causes inaccuracies. This challenge makes it tough to obtain depth maps of transparent objects and hinders the application of most robotic grasp and manipulation methods. Currently, researchers primarily explore single-view RGB-D based depth completion methods \cite{1ClearGrasp,3TransCG,4FDCT,10Rgb-d,2TranspareNet} or multi-view depth estimation methods \cite{5CDP,6MVTrans,12zhang2023multi,7Dex-NeRF,8Evo-NeRF,13wang2021nerf}. However, both methods have their own limitations. 

The single-view RGB-D based depth completion methods primarily recover object depth from the raw depth map, and the role of the RGB image is to help the network eliminate noise from the raw depth map. Most of these methods \cite{1ClearGrasp,3TransCG,4FDCT} treat the depth map as a 2D image which lack the capability to accurately perceive the 3D geometric structure of transparent objects. Remaining methods \cite{10Rgb-d,2TranspareNet}, which transform depth maps into point clouds or voxels, are characterized by multi-step pipelines, not end to end. As all of these methods are based on a single view of the RGB-D image, it is hard to gather enough spatial information from just one viewpoint in complex scenarios with heavy occlusion. This makes it tough to accurately estimate the spatial relationships among multiple transparent objects. As a result, the restored depth map often inaccurately represents overlapping objects with identical depth values.

Compared to single-view RGB-D based depth completion methods, multi-view depth estimation methods can capture richer spatial information and better understand complex spatial relationships. However, some methods \cite{5CDP,6MVTrans} utilize Multi-View Stereo (MVS) \cite{12zhang2023multi}, which are particularly susceptible to interference from the unique visual properties of transparent objects. As light can pass through transparent objects, the RGB images captured by cameras may include visible features from behind these objects, causing the images distorted. This often leads MVS methods to mistakenly bypass transparent objects and estimate the depth of the background. Such errors can result in a global shift in the depth of the object. Moreover, MVS does not capture the information from the raw depth maps which hold valuable depth information that should not be ignored and underperforms in areas with sparse textures, where feature extraction and matching are challenging. Other methods \cite{7Dex-NeRF,8Evo-NeRF} based on Neural Radiance Fields (NeRF) \cite{13wang2021nerf} require individual training for each scenario, potentially limiting their generalization in dynamic environments.

To address the limitations of both methods, we propose an end-to-end network that combines the best of single-view RGB-D based depth completion and multi-view depth estimation for the transparent object depth completion. Although the unique visual properties of transparent objects can cause partial distortion in solely RGB or depth images, the multi-view RGB images and depth maps can complement each other, effectively reducing the impact of these distortions on the network. Our single-view depth completion module primarily predicts depth from the single-view raw depth maps which performs effectively in sparsely textured areas and near object edges, compensating for the shortcomings of the multi-view depth estimation in these areas. Conversely, our multi-view depth estimation module utilizes multi-view RGB images to construct a Cost Volume that contains the spatial structural information of the objects. Multi-view RGB images not only help the network overcome the poor performance of single-view depth completion in heavily occluded scenarios but also alleviate the interference of noise in the raw depth maps. Next, we inject the depth map predicted by the single-view module into the multi-view module, correcting the global shift in depth within the Cost Volume produced by the multi-view depth estimation. Finally, we evaluate the accuracy of the depth predicted by the network in different modules and integrate these depth maps based on the pixel-level confidence maps. The experimental results demonstrate that depth completion can be achieved accurately in the transparent objects with strong robustness and generalization, even in complex scenarios with heavy occlusion and unseen scenarios, and our method has achieved the state-of-the-art performance on the ClearPose \cite{19chen2022clearpose} and TransCG \cite{3TransCG} datasets.

The contributions of this work can be summarized as follows:

$\bullet$ We propose an end-to-end network for transparent object depth completion that combines the strengths of single-view RGB-D based depth completion and multi-view depth estimation. It effectively enhances the accuracy and robustness of depth completion in complex scenarios with heavy occlusion.

$\bullet$ We introduce a depth injection module that injects depth predicted from single-view RGB-D into the cost volume generated by the plane sweep stereo of multi-view RGB images. This significantly reduces the global bias of the cost volume and effectively  enhances the accuracy of the object positions in the 3D space.

$\bullet$ We introduce a depth refinement module based on confidence estimation. This module selectively fuses the best-performing parts from multiple depth maps which further refines the restored depth map.

\section{Related Work}
\label{gen_inst}

In this section, we briefly review articles on depth completion and estimation for transparent objects. The unique visual properties of transparent objects prevent optical sensors from accurately capturing their depth maps, hindering the perception of these objects. To recover the depth of transparent objects, existing methods can be approximately divided into two categories: single-view RGB-D based depth completion and multi-view depth estimation.
\subsection{Single-View RGB-D based Transparent Object Depth Completion}

The single-view RGB-D based depth completion methods for transparent objects recover depth from a single RGB-D image. Based on the different approaches on processing the depth map, they are classified into 2D image-based methods and 3D data-based methods.

2D image-based methods \cite{1ClearGrasp,3TransCG,4FDCT,11zhang2018deep,40NEURIPS2019_e2c61965,9DepthGrasp} treat the depth as a 2D image, using convolutional neural networks to predict depth map from an RGB-D image. Zhang et al. \cite{11zhang2018deep} introduce a two-stage pipeline that uses fully convolutional neural networks and global optimization to complete depth maps. Zhong et al. \cite{40NEURIPS2019_e2c61965} use the correlation between depth maps and RGB images to perform sparse depth completion. ClearGrasp \cite{1ClearGrasp} is the pioneering work in depth completion for transparent objects. It incorporates a transparent object segmentation module into the network, which improves the network's performance on transparent objects. DepthGrasp \cite{9DepthGrasp} introduces a self-attention \cite{33vaswani2017attention} module, allowing the model to focus on areas related to transparent objects. Moreover, it replaces the global optimization in ClearGrasp with a generative adversarial network \cite{34goodfellow2014generative}, enabling end-to-end training. TransCG \cite{3TransCG} adopts the U-Net \cite{22ronneberger2015u} architecture and offers both faster processing speed and improved accuracy. FDCT \cite{4FDCT} improves TransCG by adding a fusion branch for better accuracy. These 2D image-based methods can quickly recover object depth from single RGB-D images, but they lack the ability to precisely perceive the 3D geometric structure of transparent objects. The 3D geometric structure in completed depth maps is less accurate and poor in areas with significant object shape variation, such as in overlapping objects. 3D data-based methods \cite{10Rgb-d,2TranspareNet} first reconstruct objects in 3D space, then convert the 3D data into depth maps. LIDF \cite{10Rgb-d} estimates transparent object depth by learning a local implicit function for ray-voxel interactions. Xu et al. propose TranspareNet \cite{2TranspareNet}, a two-stage pipeline that combines point cloud completion and depth completion. These 3D data-based methods struggle in scenarios with heavy occlusion when using single-view images, as they do not provide sufficient information for accurate transparent objects reconstruction. Additionally, these methods cannot be trained end-to-end.

\subsection{Multi-View Transparent Object Depth Estimation}

Multi-view depth estimation for transparent objects skips the unreliable raw depth data from sensors, instead estimating depth from multi-view RGB images. Yao et al. introduce MVS-Net \cite{12zhang2023multi}, a deep learning architecture that utilizes homography warping for dense depth map inference from multi-view images. Some methods \cite{41NEURIPS2022_38e511a6,42NEURIPS2022_94ef7217,16zhang2023geomvsnet} improve MVS to enhance the accuracy of depth estimation, but they are not suitable for transparent objects. Wang et al. \cite{6MVTrans} utilize MVS to perceive transparent objects, directly obtaining depth maps, 3D bounding boxes, and poses from multi-view RGB images. Cai et al. \cite{5CDP} utilize PointNet \cite{35qi2017pointnet} to extract depth information from background point clouds, which provides absolute scale to MVS and ensures consistent depth prediction structures. These methods aim to estimate depth with more precise spatial structures. However, low-textured, specular, and reflective regions on transparent objects make dense matching challenging, thereby affecting the accuracy of depth estimation. Furthermore, these methods exhibit a global bias in their depth estimates because they lack accurate object depth guidance. Mildenhall et al. propose Neural Radiance Fields (NeRF)  \cite{13wang2021nerf}, a neural network that learns from multi-view images to represent reconstructed scenarios. Jeffrey et al. propose Dex-NeRF  \cite{7Dex-NeRF} and Evo-NeRF  \cite{8Evo-NeRF}, methods using NeRF to represent non-Lambertian effects such as specularities and reflections, and to estimate the depth of transparent objects. However, the inherent limitations of NeRF in generalization and precise 3D structure reconstruction can compromise the effectiveness of these NeRF-based methods.
 
 To avoid these problems, our method integrates single-view RGB-D based depth completion with multi-view depth estimation. Our single-view depth completion module completes a depth map from a single RGB-D image, which exhibits accurate global depth. This module typically performs well in areas with low texture and at the edges of objects. Our multi-view module estimates a depth map from multi-view RGB images, in which the objects exhibit accurate 3D geometric shapes. This module is capable of estimating differences in relative object depth and can effectively handle complex scenarios with significant object occlusions. We effectively fuse the features of both modules and retain their individual strengths through depth infusion and confidence estimation. We demonstrate that our method surpasses the two existing categories of methods.
 
\begin{figure*}
	\centering
		\includegraphics[width=1.0\linewidth]{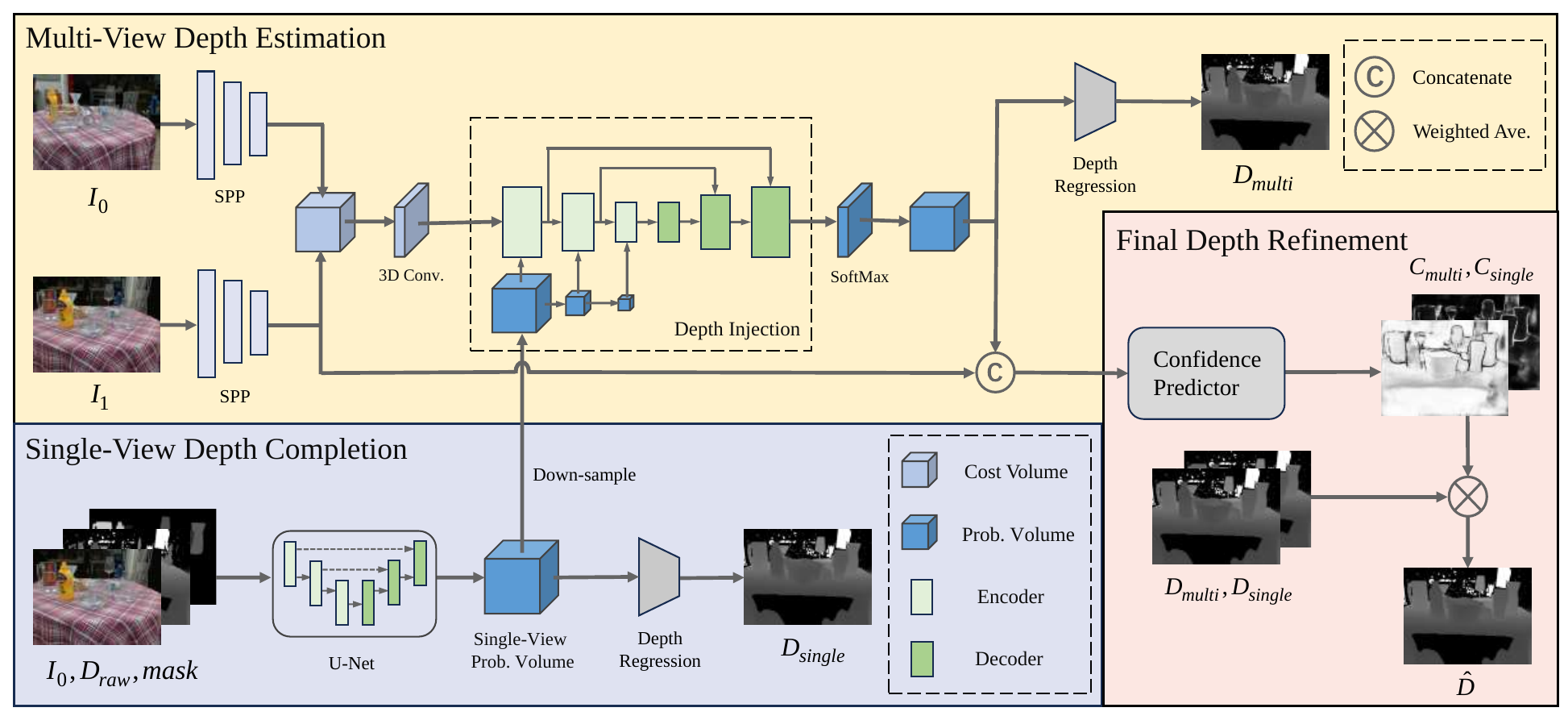}
	\caption{Overview of our proposed method. We predict the depth maps separately from a single RGB-D image and multi-view RGB images. Then, the single-view probability volume is injected into the multi-view depth estimation. Finally, we predict the confidence of these depth maps and refine the restored depth.}
	\label{fig:1overview}
 % \vspace{-8pt}
\end{figure*}

\section{Method}
\label{headings}

In this section, we present an end-to-end transparent object depth completion network, illustrated in Figure \ref{fig:1overview}. We aim to predict the restored depth $\hat{D} \in \mathbb{R}^{H \times W}$ for transparent objects from given inaccurate raw depth map $D_{raw} \in \mathbb{R}^{H \times W}$, two RGB images $I_{0}, I_{1} \in \mathbb{R}^{H \times W \times 3}$ from different viewpoints, and a mask of the transparent objects. Our network consists of three parts: single-view depth completion, multi-view depth estimation, and final depth refinement. In the single-view depth completion module, single-view depth $D_{single} \in \mathbb{R}^{H \times W }$ is calculated from $D_{raw}$, $I_{0}$, and the mask of the transparent objects. In the multi-view depth estimation module, the multi-view cost volume is generated by plane sweep stereo from $I_{0}, I_{1}$. Then, the multi-view depth $D_{multi} \in \mathbb{R}^{H \times W}$ is estimated from this cost volume. The details of these two modules can be found in Section \ref{subsec:Asingle_view_depth}. Finally, we predict confidence maps $C_{multi}$, $C_{single} \in \mathbb{R}^{H \times W}$ for depth maps and compute the restored depth $\hat{D}$ based on them through the final depth refinement module. The details can be found in Section \ref{subsec:DFinal Depth Refinement}.

\subsection{Single-View Depth Completion and Multi-View Depth Estimation}

\label{subsec:Asingle_view_depth}
$\textbf{Single-View Depth Completion:}$
% \subsubsection{Single-view depth completion}
Inspired by TransCG \cite{3TransCG}, we introduce a U-Net network to reconstruct transparent objects from the raw depth $D_{raw}$, the image $I_{0}$, and the mask. To enhance the perception of spatial information, we divide the depth space into $N$ hypothesis depth planes at equal intervals. Then, the U-Net predicts the single-view probability volume $P$, which represents the probability that each pixel attaches to a specific depth plane. Finally, we utilize a depth regression module \cite{17kendall2017end} to predict the single-view depth $D_{single}$. This module calculates $D_{single}$ through a weighted summation of the probability volume:
\begin{equation}
\begin{split}
D_{single}=\sum_{i=0}^{N}d_{i}\times P(i)
\end{split}
\end{equation}
 
where $P(i)$ is the probability of points with a depth of $d_{i}$.

\begin{figure*}
	\centering
		\includegraphics[width=0.9\linewidth]{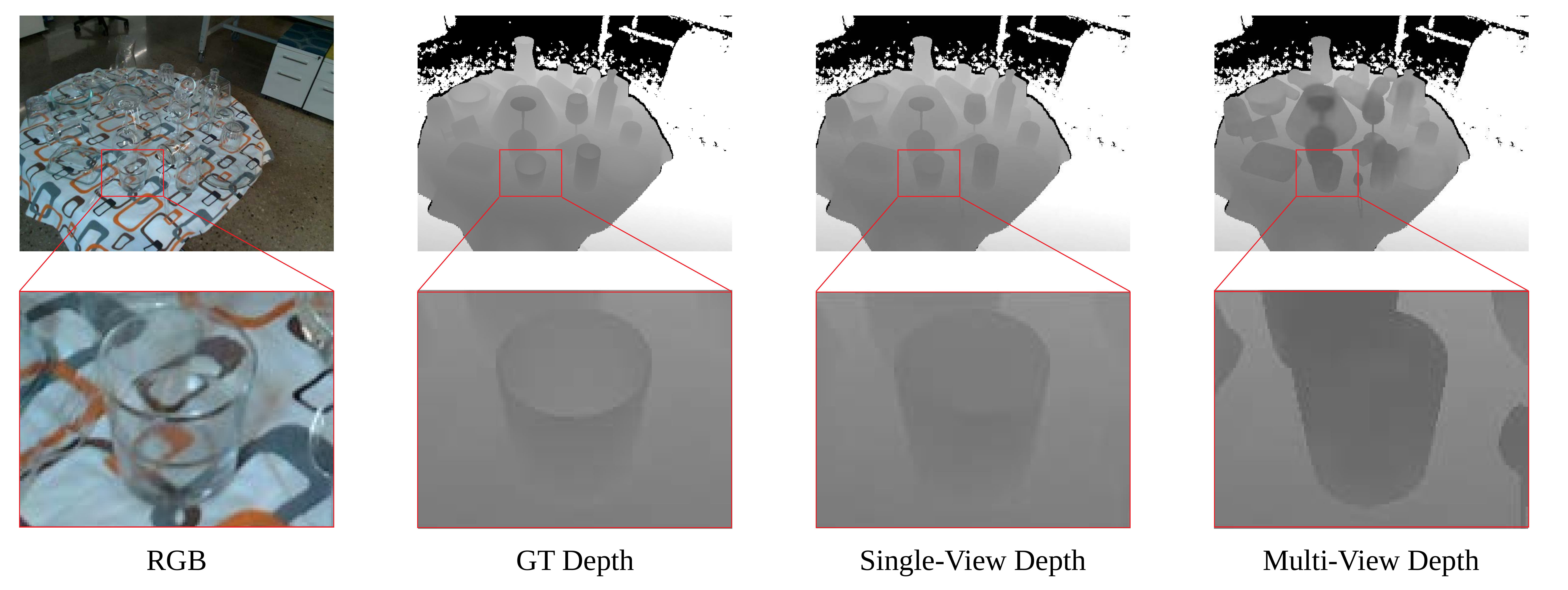}
	\caption{Visualization of single-view and multi-view depth. The second row shows an enlarged view of the depth map. In the depth maps, the lighter gray value indicates a greater depth value.}
	\label{fig:5method}
% \vspace{-8pt}
\end{figure*}

The design of the Single-View Depth Completion module is based on two premises: (1) Although there are gaps and inaccuracies in the raw depth $D_{raw}$, optical sensors can still capture correct depth values for certain parts of objects. (2) The size of transparent objects is relatively small compared to the entire scenario, so the distribution of actual depth within the objects is relatively concentrated. The single-view module treats accurate depth values as reference points. Since the depth values for the same object do not vary significantly, this module can accurately predict the global depth of the entire object. As shown in the third column of Figure \ref{fig:5method}, the gray values representing the depth of the cup in the single-view depth are close to the ground truth. The accurate global depth of the single-view depth $D_{single}$ effectively reduces network errors and serves as a reference for other modules. Additionally, the network input mask helps the network focus on transparent objects and provides more stability during training.

$\textbf{Multi-View Depth Estimation:}$
% \subsubsection{Multi-view depth estimation}
Within this module, multi-view depth $D_{multi}$ is estimated from the reference image $I_{0}$ and the source image $I_{1}$. Initially, we extract features $F_{0}, F_{1}$ from both the reference image $I_{0}$ and the source image $I_{1}$ using a spatial pyramid pooling (SPP) \cite{21he2015spatial} module with shared parameters. Inspired by MVS-Net \cite{12zhang2023multi}, features $F_{0}$ and $F_{1}$ are warped into previously defined depth planes of the reference camera to form feature volumes. Subsequently, these feature volumes are aggregated into the multi-view cost volume. To eliminate noise caused by light reflecting and refracting from transparent objects, the multi-view cost volume is refined with 3D convolutions. Although this cost volume contains rich spatial geometric information, the global depth of objects within them is imprecise. As shown in the fourth column of Figure \ref{fig:5method}, the gray values representing the depth of the cup in the multi-view depth exhibit a significant bias relative to the ground truth. Indeed, the depth estimation based on disparity tends to have limited accuracy due to the lack of direct depth guidance. 

To address this issue, we introduce the depth injection module. Inspired by GEO-MVS \cite{16zhang2023geomvsnet}, a network composed of multiple encoders and decoders is designed to fuse the single-view probability volume into the multi-view cost volume at various scales. The initial encoder of this network fuses the multi-view cost volume and the single-view probability volume, outputting an encoded cost volume at a reduced scale. Then, subsequent encoders receive progressively encoded cost volumes concatenated with single-view probability volumes, which match the dimensions of the corresponding cost volume through 3D MaxPooling. These encoded outputs are processed through decoders that output an enhanced multi-view cost volume, embedding the depth information from the single-view module. The depth from the single-view module provides global perception of transparent objects in the scenario for the multi-view module, correcting the depth offset of the entire transparent object in the cost volume. Moreover, the rich depth information and accurate background depth guide the cost volume in scenario reconstruction, improving the robustness and accuracy of cost matching.

Following this process, a normalized probability volume is generated from the enhanced multi-view cost volume through a softmax layer. Finally, the multi-view depth $D_{multi}$ is estimated from the multi-view probability volume.

\subsection{Final Depth Refinement}
\label{subsec:DFinal Depth Refinement}

Although the bias of multi-view module is reduced through the multi-scale depth injection, the single-view depth completion and the multi-view depth estimation exhibit differences in different object areas due to their differences in input and network structure. Specifically, the single-view module achieves higher accuracy near object edges, as it can learn the contours of objects from RGB images and masks. While the multi-view module demonstrates enhanced performance in areas with complex geometries due to its stronger ability to perceive the 3D geometric structure of transparent objects from varied perspectives. To further fuse these depth maps, we introduce a depth refinement module based on confidence estimation.

We estimate pixel-level confidence maps $C_{multi}$, $C_{single}$ for multi-view depth $D_{multi}$ and single-view depth $D_{single}$ using reference image features $F_{0}$ and the multi-view probability volume. Then, the confidence map is used as weights to calculate the weighted average of their respective depths. The restored depth $\hat{D}$ can be computed as follows:
\begin{equation}
\begin{split}
\hat{D}=C_{multi}\otimes D_{multi}+C_{single}\otimes D_{single}
\end{split}
\end{equation}
where  $\otimes$ represents elementwise multiplication.

\subsection{Loss Function}

Our network is trained using the following loss function:

 \begin{equation}
\begin{split}
 L=L_{restored}+\lambda_{1}L_{multi}+\lambda_{2}L_{single}+\lambda_{3}L_{normal}
\end{split}
\end{equation}

where $\lambda_{1}$, $\lambda_{2}$ and $\lambda_{3}$ are the weight parameters. $L_{restored}$, $L_{multi}$, and $L_{single}$ respectively represent the $L_1$ losses for the restored depth $\hat{D}$, the multi-view depth $D_{multi}$, and the single-view depth $D_{single}$ against the ground truth depth. $L_{normal}$ is the $L_1$ loss between the surface normals computed from the single-view depth $D_{single}$ and the ground truth depth respectively, which helps mitigate the limitations of the single-view module in learning the 3D geometric shapes of objects. To ensure the network focuses on transparent objects, we compute all loss functions within the masked region of the transparent objects.

All our experiments are conducted using the Adam optimizer \cite{20kingma2014adam} with an initial learning rate of 0.0001 and a batch size of 4. The weights $\lambda_{1}$, $\lambda_{2}$ and $\lambda_{3}$ for the loss function are set at 0.8, 0.5, and 0.0005, respectively. The number of hypothesis planes $N$ is set to 45. For ClearPose, images are resized to a resolution of 640 × 480 and trained for 4 epochs, with an interval of 40 frames between the reference image and the source image. For TransCG, to maintain consistency with other methods \cite{3TransCG,4FDCT}, images are resized to a resolution of 320 × 240 and trained for 20 epochs. Images captured by the D435 camera are used as reference frames, while those from the L515 camera serve as source images. Both training and testing utilize an NVIDIA GeForce RTX 4090 GPU.
\vspace{-5pt}

\section{Experiment}

\label{others}

\subsection{Datasets and Metrics}
$\textbf{Datasets}$:
% \subsubsection{Dataset}
For our study, we use two datasets: TransCG \cite{3TransCG} and ClearPose \cite{19chen2022clearpose}. TransCG represents the first large-scale, real-world dataset that provides ground truth depth, surface normals, and transparent masks in diverse and cluttered scenarios. ClearPose is a large-scale transparent object dataset, containing over 350,000 labeled real-world RGB-Depth frames and 5 million instance annotations covering 63 household objects.  

$\textbf{Metrics}$:
To assess the accuracy of our depth completion model, we utilize a set of standard evaluation metrics. These include the Root Mean Squared Error (RMSE), the Absolute Relative Difference (REL), the Mean Absolute Error (MAE), and a set of Threshold values at 1.05, 1.10, and 1.25 levels, following \cite{1ClearGrasp}. It is important to mention that all metrics are calculated on the masks of transparent objects, with RMSE and MAE computed in meters.

\subsection{Comparison with the State-of-the-art Methods}
\begin{table*}
    \caption{Quantitative comparison to the state-of-the-art methods}
    \label{table:5depth_completion_detailed}
    \centering
    \footnotesize
    \begin{tabular*}{\textwidth}{@{\extracolsep{\fill}} c|c|c c c c c c}
    \toprule
    Testset & Method & RMSE↓ & REL↓ & MAE↓ & $\delta1.05$ ↑ & $\delta1.10$ ↑ & $\delta1.25$ ↑ \\ 
    \toprule
    \multicolumn{8}{c}{Part1: Training on the entire ClearPose dataset}\\
    \toprule
   \multirow{3}{*}{New Background} 
                            & CDPTR* & 0.1252 & 0.1241 & 0.1048 & 26.26 & 48.51 & 86.37 \\
                            & TranspareNet* & 0.0290 & \textbf{0.0182} & 0.0163 & 92.90 & 97.96 & 99.62 \\
                            & Ours & \textbf{0.0222} & 0.0183 & \textbf{0.0159} & \textbf{94.50} & \textbf{99.04} & \textbf{99.95} \\

    \hline
    \multirow{3}{*}{Heavy Occlusion} 
                                     & CDPTR* & 0.1408 & 0.1195 & 0.1537 & 20.97 & 40.70 & 79.76 \\
                                     & TranspareNet* & 0.0539 & 0.0340 & 0.0320 & 80.18 & 92.19 & 98.67 \\
                                     & Ours & \textbf{0.0343} & \textbf{0.0257} & \textbf{0.0230} & \textbf{87.01} & \textbf{96.97} & \textbf{99.75} \\
    \hline
    \multirow{3}{*}{Translucent Cover} 
                                   & CDPTR* & 0.1563 & 0.1357 & 0.1564 & 22.49 & 40.36 & 72.71 \\
                                   & TranspareNet* & 0.1204 & 0.1118 & 0.0961 & 30.04 & 52.20 & 84.07 \\
                                   & Ours & \textbf{0.0876} & \textbf{0.0798} & \textbf{0.0690} & \textbf{43.78} & \textbf{68.53} & \textbf{92.75} \\
                                   
    \hline
    \multirow{3}{*}{Opaque Distractor} 
                                       & CDPTR* & 0.0972 & 0.0824 & 0.0983 & 32.04 & 57.89 & 94.43 \\
                                       & TranspareNet* & 0.0755 & 0.0613 & 0.0522 & 55.68 & 75.79 & 97.59 \\
                                       & Ours & \textbf{0.0570} & \textbf{0.0502} & \textbf{0.0415} & \textbf{63.11} & \textbf{87.13} & \textbf{97.65} \\                                    
    \hline
    \multirow{3}{*}{Filled Liquid} 
                                   & CDPTR* & 0.1198 & 0.1296 & 0.1001 & 31.34 & 54.21 & 84.86 \\
                                   & TranspareNet* & 0.0495 & 0.0367 & 0.0314 & 77.43 & 92.01 & 98.87 \\
                                   & Ours & \textbf{0.0363} & \textbf{0.0302} & \textbf{0.0253} & \textbf{81.36} & \textbf{96.38} & \textbf{99.72} \\
    \hline

    \multirow{3}{*}{Non Planar} 
                            & CDPTR* & 0.1709 & 0.1428 & 0.1514 & 22.62 & 41.17 & 74.44 \\
                            & TranspareNet* & 0.0854 & 0.0618 & 0.0583 & 61.68 & 81.45 & 95.42 \\
                            & Ours & \textbf{0.0663} & \textbf{0.0456} & \textbf{0.0437} & \textbf{72.81} & \textbf{89.66} & \textbf{97.12} \\

    \toprule
    \multicolumn{8}{c}{Part2: Training on the set 1 of ClearPose dataset}\\
    \toprule
    \multirow{3}{*}{Set 1} &TransCG & 0.048 & 0.038 & 0.033 & 76.36 & 94.22 & 99.40 \\
                            &FDCT & 0.045 & 0.033 & 0.028 & 82.15 & 94.43 & 99.25 \\
                            &Ours & \textbf{0.034} & \textbf{0.024} & \textbf{0.021} & \textbf{88.08} & \textbf{95.32} & \textbf{99.77} \\
    \toprule
    \multicolumn{8}{c}{Part3: Training on the TransCG dataset}\\
    \toprule
    \multirow{5}{*}{TransCG} & ClearGrasp & 0.054 & 0.083 & 0.037 & 50.48 & 68.68 & 95.28 \\
                                &LIDF & 0.019 & 0.034 & 0.015 & 78.22 & 94.26 & 99.80 \\
                                &TransCG & 0.018 & 0.027 & 0.012 & 83.76 & 95.67 & 99.71 \\  
                                &FDCT & 0.015 & 0.022 & 0.010 & 88.18 & 97.15 & 99.81 \\
                                &Ours & \textbf{0.010} & \textbf{0.014} & \textbf{0.006} & \textbf{94.64} & \textbf{98.83} & \textbf{99.97} \\
    \bottomrule
    \end{tabular*}
    \par % 添加段落换行
    \scriptsize
    \raggedright
    \textbf{Note:} The * marks the result reproduced by us.
    \vspace{-6pt}

\end{table*}

 First, we compare our method on the ClearPose dataset with TranspareNet \cite{2TranspareNet} and CDPTR \cite{5CDP}. The results, as shown in the 1st part of Table \ref{table:5depth_completion_detailed}, show that our method performs the best across nearly all metrics in all scenarios. The improvement in the Translucent Cover scenario of the ClearPose dataset is particularly significant, with a decrease of 28.2$\%$(0.0961 vs. 0.0690) in the MAE metric compared to TranspareNet. This significant advancement primarily results from reduced sensitivity to noise in our method. Additionally, considering that FDCT \cite{4FDCT} and TransCG are only trained and tested on set 1 of the ClearPose dataset, we conduct fair comparison by testing our method under the same conditions \cite{4FDCT}. The results in the 2nd part of Table \ref{table:5depth_completion_detailed} also demonstrate strong performance across all metrics. Particularly, compared to FDCT, our method achieves a 25.0$\%$(0.028 vs.0.021) reduction in the MAE metric. To further confirm the effectiveness of our method, we train and test on TransCG dataset against TransCG, FDCT, ClearGrasp \cite{1ClearGrasp} and LIDF \cite{10Rgb-d}. As shown in the 3rd part of Table \ref{table:5depth_completion_detailed}, our method maintains consistent superiority across all metrics. Compared to FDCT, we achieve a 40.0$\%$(0.010 vs. 0.006) reduction in the MAE metric. 

\begin{figure*}
	\centering
		\includegraphics[width=0.95\linewidth]{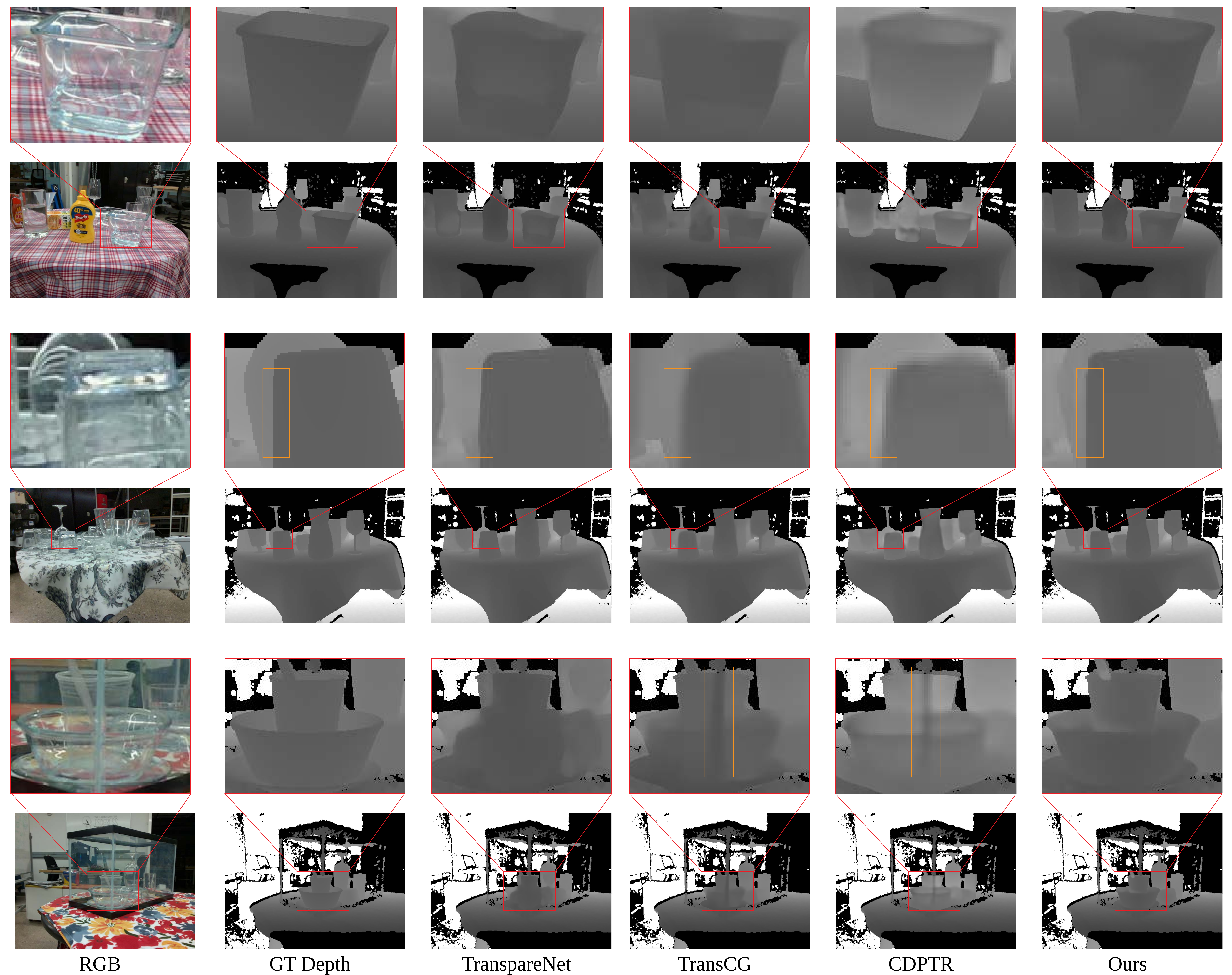}
	\caption{ Qualitative comparison to the state-of-the-art methods.}
	\label{fig:4Qualitative}
\vspace{-8pt}
\end{figure*}

To further demonstrate the advantages of our method, the qualitative results of depth completion and estimation on the ClearPose dataset are shown in Figure \ref{fig:4Qualitative}. The 1st row shows the enlarged depth maps for the transparent object without occlusion. Our method more accurately reconstructs the geometric structure of the container's upper edge compared to single-view methods like TranspreNet and TransCG. Additionally, compared to the ground truth, the gray values of the container reconstructed by CDPTR appear lighter, whereas our method produces gray values that are closer to the ground truth. The 3rd row shows the depth maps for transparent objects with occlusion. Compared to other methods, our method displays more obvious edges, as shown in the yellow box. The 5th row shows the depth maps for transparent objects placed inside a translucent box. TransCG and CDPTR mistakenly reconstruct the edges of the translucent box in front of a cup and bowl shown in the yellow box, whereas our completed depth map does not include the translucent box, consistent with the ground truth. These results show that our method is more accurate and robust.

\subsection{Ablation Studies}
% Compared to TranspareNet, our method shows sharper edges within the yellow box.
To demonstrate the effectiveness of each module of our method, we design two experiments to validate the effectiveness of the depth injection module and the depth refinement module respectively on the ClearPose dataset across six scenarios and average the results.

\subsubsection{Ablation Study on the Depth Injection}

\begin{figure*}
	\centering
		\includegraphics[width=0.85\linewidth]{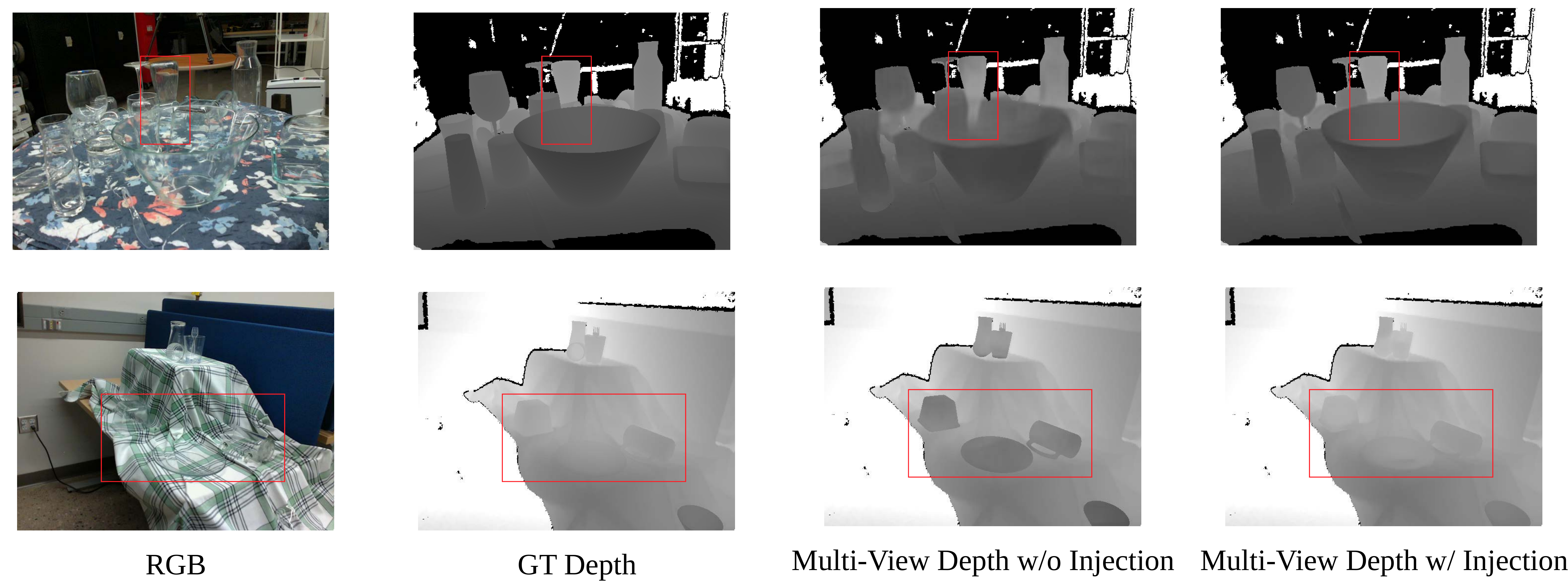}
	\caption{Qualitative results of depth injection.}
	\label{fig:2inject}
\vspace{-12pt}
\end{figure*}

To evaluate the impact of depth injection, we design a baseline by replacing the depth injection module with additional 3D convolutions in our network. The qualitative results are shown in Figure \ref{fig:2inject}. In the 1st image of the 3rd column, we observe that the multi-view depth estimation module mistakenly bypasses the bowl and estimates the depth of the cup indicated by the red box. The 2nd image in the 3rd column shows that the gray values of the objects in the red box are darker, indicating that their depth values are smaller than the ground truth. As shown in the fourth column, the injection module effectively corrects the errors in the previously described multi-view depth. Then, we quantitatively analyze the effectiveness of the depth injection module on multi-view depth, as shown in Table \ref{table:3depth_injection}. All metrics clearly demonstrate that depth injection enhances performance compared to no injection. Specifically, the depth injection module reduces the MAE metric by 64.1\%(0.1046vs.0.0375). This further demonstrates the effectiveness of the depth injection module.
\begin{table}
    \vspace{-4pt}

    \caption{Quantitative results of the depth injection on the multi-view depth}
    \label{table:3depth_injection}
    \centering
    \footnotesize
    \setlength{\tabcolsep}{11pt}
    \begin{tabular*}{\textwidth}{@{\extracolsep{\fill}}c| c c c c c c}
    \toprule
    Method & RMSE↓ & REL↓ & MAE↓ & $\delta1.05$ ↑ & $\delta1.10$ ↑ & $\delta1.25$ ↑ \\ 
    \toprule
    w/o Injection & 0.1254 & 0.1161 & 0.1046 & 31.23 & 54.43 & 84.81 \\
    w/ Injection & \textbf{0.0518} & \textbf{0.0430} & \textbf{0.0375} & \textbf{72.56} & \textbf{89.33} & \textbf{97.86} \\
    \bottomrule
    \end{tabular*}
    \vspace{-4pt}
\end{table}

\begin{figure*}
	\centering
		\includegraphics[width=0.85\linewidth]{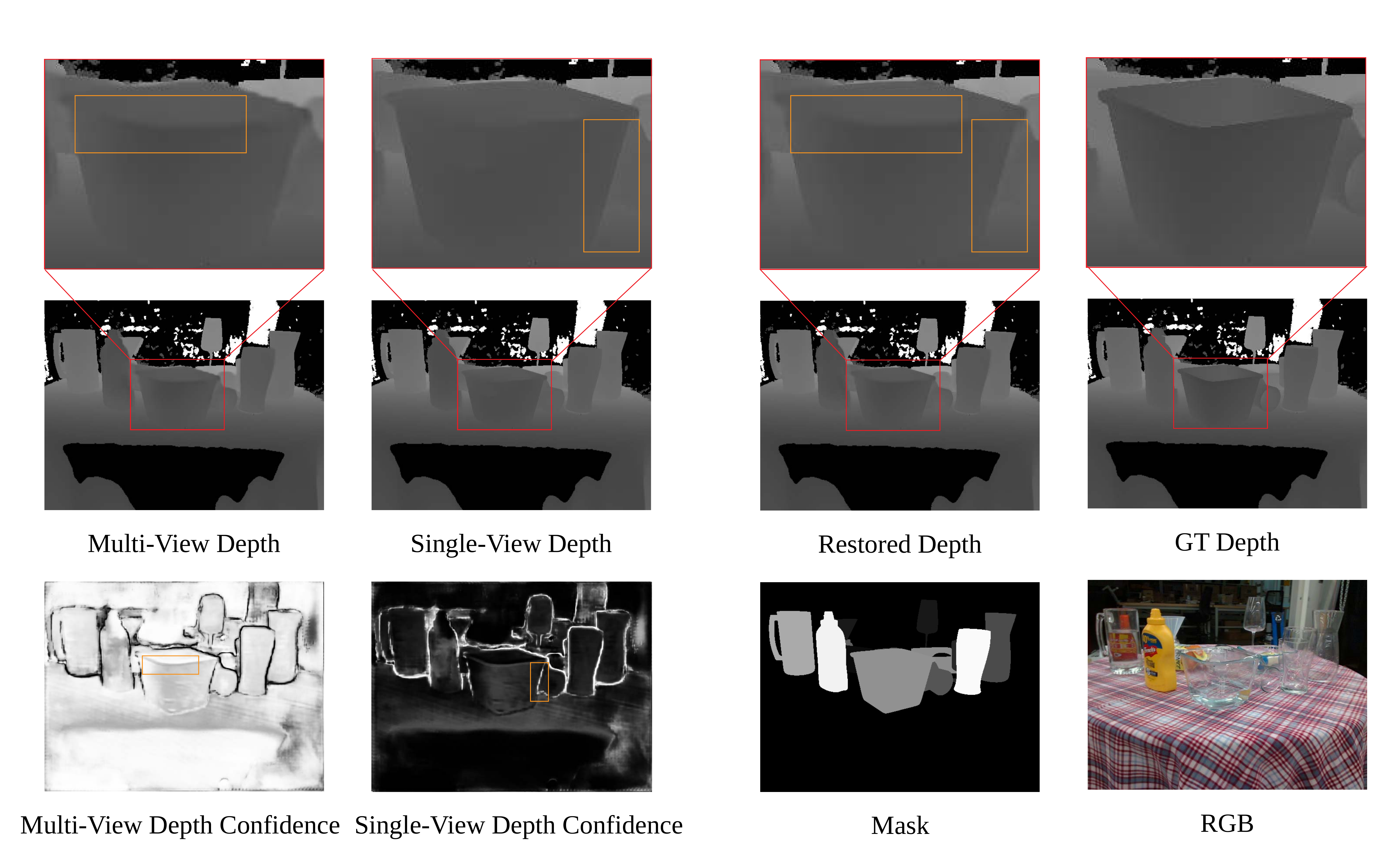}
	\caption{Qualitative results of the depth refinement. In the confidence maps, the brighter gray value indicates a higher confidence.}
	\label{fig:3Confidence}
 % \vspace{-8pt}
\end{figure*}
\begin{table}
    \caption{Quantitative results of the depth refinement}
    \label{table:4depth_refinement}
    \centering
    \footnotesize
    \setlength{\tabcolsep}{11pt}
    \begin{tabular*}{\textwidth}{@{\extracolsep{\fill}}c| c c c c c c}
    \toprule
    Method & RMSE↓ & REL↓ & MAE↓ & $\delta1.05$ ↑ & $\delta1.10$ ↑ & $\delta1.25$ ↑ \\ 
    \toprule
    w/o Refinement & 0.0518 & 0.0430 & 0.0375 & 72.56 & 89.33 & \textbf{97.86} \\
    w/ Refinement & \textbf{0.0506} & \textbf{0.0416} & \textbf{0.0364} & \textbf{73.76} & \textbf{89.62} &97.82 \\
    \bottomrule
    \end{tabular*}
\end{table}

\subsubsection{Ablation Study on the Depth Refinement}
 
To demonstrate the role of the refinement module in depth completion, we present a multi-view depth map, a single-view depth map and their weighted average with their confidence maps as weights from left to right, as shown in the first 2 rows of Figure \ref{fig:3Confidence}. From their enlarged parts of the container shown in the yellow boxes, it can be observed that the multi-view depth map presents more detailed structure of a transparent object from varied perspectives while the single-view depth map displays a sharper edge.Their weighted average (refined depth map) precisely combines the advantages of both to yield the result closest to the ground true. From the confidence maps in the 3rd row, we can also see that gray values in two the yellow boxes respectively corresponding to the detailed structure and the sharper edge are much brighter.

We further conduct a quantitative experiment to compare the refined depth map with the multi-view depth map containing more details. It can be seen from table \ref{table:4depth_refinement} that the depth refinement module improves performance in five out of six metrics which once again proves the effectiveness of the depth refinement module.

\section{Conclusion}
In this work, we propose an end-to-end network for transparent object depth completion that combines the best of single-view RGB-D based depth completion and multi-view depth estimation. The depth injection module enhances the accuracy of object positioning in 3D space, while our depth refinement module improves the accuracy and robustness of depth completion. The experiments show that our method provides more accurate estimations of geometric structures compared to single-view methods, and more accurate global depth compared to multi-view methods. Our method surpasses the state-of-the-art methods on the TransCG and ClearPose datasets. Although our method achieves significant improvement, it still has a limitation. Compared to some single-view RGB-D based depth completion methods, our method has higher GPU memory occupation.

\bibliographystyle{IEEEtran}

\bibliography{IEEEabrv,IEEEexample}

\begin{thebibliography}{10}
\providecommand{\url}[1]{#1}
\csname url@rmstyle\endcsname
\providecommand{\newblock}{\relax}
\providecommand{\bibinfo}[2]{#2}
\providecommand\BIBentrySTDinterwordspacing{\spaceskip=0pt\relax}
\providecommand\BIBentryALTinterwordstretchfactor{4}
\providecommand\BIBentryALTinterwordspacing{\spaceskip=\fontdimen2\font plus
\BIBentryALTinterwordstretchfactor\fontdimen3\font minus \fontdimen4\font\relax}
\providecommand\BIBforeignlanguage[2]{{%
\expandafter\ifx\csname l@#1\endcsname\relax
\typeout{** WARNING: IEEEtran.bst: No hyphenation pattern has been}%
\typeout{** loaded for the language `#1'. Using the pattern for}%
\typeout{** the default language instead.}%
\else
\language=\csname l@#1\endcsname
\fi
#2}}

\bibitem{27chen2023efficient}
S.~Chen, W.~Tang, P.~Xie, W.~Yang, and G.~Wang, ``Efficient heatmap-guided 6-dof grasp detection in cluttered scenes,'' \emph{IEEE Robotics and Automation Letters}, 2023.

\bibitem{28barad2023graspldm}
K.~R. Barad, A.~Orsula, A.~Richard, J.~Dentler, M.~Olivares-Mendez, and C.~Martinez, ``Graspldm: Generative 6-dof grasp synthesis using latent diffusion models,'' \emph{arXiv preprint arXiv:2312.11243}, 2023.

\bibitem{29tang2023graspgpt}
C.~Tang, D.~Huang, W.~Ge, W.~Liu, and H.~Zhang, ``Graspgpt: Leveraging semantic knowledge from a large language model for task-oriented grasping,'' \emph{IEEE Robotics and Automation Letters}, 2023.

\bibitem{1ClearGrasp}
S.~Sajjan, M.~Moore, M.~Pan, G.~Nagaraja, J.~Lee, A.~Zeng, and S.~Song, ``Clear grasp: 3d shape estimation of transparent objects for manipulation,'' in \emph{2020 IEEE International Conference on Robotics and Automation (ICRA)}, 2020, pp. 3634--3642.

\bibitem{3TransCG}
H.~Fang, H.-S. Fang, S.~Xu, and C.~Lu, ``Transcg: A large-scale real-world dataset for transparent object depth completion and a grasping baseline,'' \emph{IEEE Robotics and Automation Letters}, vol.~7, no.~3, pp. 7383--7390, 2022.

\bibitem{4FDCT}
T.~Li, Z.~Chen, H.~Liu, and C.~Wang, ``Fdct: Fast depth completion for transparent objects,'' \emph{IEEE Robotics and Automation Letters}, vol.~8, no.~9, pp. 5823--5830, 2023.

\bibitem{10Rgb-d}
L.~Zhu, A.~Mousavian, Y.~Xiang, H.~Mazhar, J.~van Eenbergen, S.~Debnath, and D.~Fox, ``Rgb-d local implicit function for depth completion of transparent objects,'' in \emph{Proceedings of the IEEE/CVF Conference on Computer Vision and Pattern Recognition}, 2021, pp. 4649--4658.

\bibitem{2TranspareNet}
\BIBentryALTinterwordspacing
H.~Xu, Y.~R. Wang, S.~Eppel, A.~Aspuru{-}Guzik, F.~Shkurti, and A.~Garg, ``Seeing glass: Joint point cloud and depth completion for transparent objects,'' \emph{CoRR}, vol. abs/2110.00087, 2021. [Online]. Available: \url{https://arxiv.org/abs/2110.00087}
\BIBentrySTDinterwordspacing

\bibitem{5CDP}
Y.~Cai, Y.~Zhu, H.~Zhang, and B.~Ren, ``Consistent depth prediction for transparent object reconstruction from rgb-d camera,'' in \emph{Proceedings of the IEEE/CVF International Conference on Computer Vision (ICCV)}, October 2023, pp. 3459--3468.

\bibitem{6MVTrans}
Y.~R. Wang, Y.~Zhao, H.~Xu, S.~Eppel, A.~Aspuru-Guzik, F.~Shkurti, and A.~Garg, ``Mvtrans: Multi-view perception of transparent objects,'' in \emph{2023 IEEE International Conference on Robotics and Automation (ICRA)}, 2023, pp. 3771--3778.

\bibitem{12zhang2023multi}
Y.~Yao, Z.~Luo, S.~Li, T.~Fang, and L.~Quan, ``Mvsnet: Depth inference for unstructured multi-view stereo,'' in \emph{Proceedings of the European conference on computer vision (ECCV)}, 2018, pp. 767--783.

\bibitem{7Dex-NeRF}
J.~Ichnowski, Y.~Avigal, J.~Kerr, and K.~Goldberg, ``Dex-nerf: Using a neural radiance field to grasp transparent objects,'' \emph{arXiv preprint arXiv:2110.14217}, 2021.

\bibitem{8Evo-NeRF}
J.~Kerr, L.~Fu, H.~Huang, Y.~Avigal, M.~Tancik, J.~Ichnowski, A.~Kanazawa, and K.~Goldberg, ``Evo-nerf: Evolving nerf for sequential robot grasping of transparent objects,'' in \emph{6th annual conference on robot learning}, 2022.

\bibitem{13wang2021nerf}
B.~Mildenhall, P.~P. Srinivasan, M.~Tancik, J.~T. Barron, R.~Ramamoorthi, and R.~Ng, ``Nerf: Representing scenes as neural radiance fields for view synthesis,'' \emph{Communications of the ACM}, vol.~65, no.~1, pp. 99--106, 2021.

\bibitem{19chen2022clearpose}
X.~Chen, H.~Zhang, Z.~Yu, A.~Opipari, and O.~Chadwicke~Jenkins, ``Clearpose: Large-scale transparent object dataset and benchmark,'' in \emph{European conference on computer vision}.\hskip 1em plus 0.5em minus 0.4em\relax Springer, 2022, pp. 381--396.

\bibitem{11zhang2018deep}
Y.~Zhang and T.~Funkhouser, ``Deep depth completion of a single rgb-d image,'' in \emph{Proceedings of the IEEE conference on computer vision and pattern recognition}, 2018, pp. 175--185.

\bibitem{40NEURIPS2019_e2c61965}
\BIBentryALTinterwordspacing
Y.~Zhong, C.-Y. Wu, S.~You, and U.~Neumann, ``Deep rgb-d canonical correlation analysis for sparse depth completion,'' in \emph{Advances in Neural Information Processing Systems}, H.~Wallach, H.~Larochelle, A.~Beygelzimer, F.~d\textquotesingle Alch\'{e}-Buc, E.~Fox, and R.~Garnett, Eds., vol.~32.\hskip 1em plus 0.5em minus 0.4em\relax Curran Associates, Inc., 2019. [Online]. Available: \url{https://proceedings.neurips.cc/paper_files/paper/2019/file/e2c61965b5e23b47b77d7c51611b6d7f-Paper.pdf}
\BIBentrySTDinterwordspacing

\bibitem{9DepthGrasp}
Y.~Tang, J.~Chen, Z.~Yang, Z.~Lin, Q.~Li, and W.~Liu, ``Depthgrasp: Depth completion of transparent objects using self-attentive adversarial network with spectral residual for grasping,'' in \emph{2021 IEEE/RSJ International Conference on Intelligent Robots and Systems (IROS)}.\hskip 1em plus 0.5em minus 0.4em\relax IEEE, 2021, pp. 5710--5716.

\bibitem{33vaswani2017attention}
A.~Vaswani, N.~Shazeer, N.~Parmar, J.~Uszkoreit, L.~Jones, A.~N. Gomez, {\L}.~Kaiser, and I.~Polosukhin, ``Attention is all you need,'' \emph{Advances in neural information processing systems}, vol.~30, 2017.

\bibitem{34goodfellow2014generative}
I.~Goodfellow, J.~Pouget-Abadie, M.~Mirza, B.~Xu, D.~Warde-Farley, S.~Ozair, A.~Courville, and Y.~Bengio, ``Generative adversarial nets,'' \emph{Advances in neural information processing systems}, vol.~27, 2014.

\bibitem{22ronneberger2015u}
O.~Ronneberger, P.~Fischer, and T.~Brox, ``U-net: Convolutional networks for biomedical image segmentation,'' in \emph{Medical image computing and computer-assisted intervention--MICCAI 2015: 18th international conference, Munich, Germany, October 5-9, 2015, proceedings, part III 18}.\hskip 1em plus 0.5em minus 0.4em\relax Springer, 2015, pp. 234--241.

\bibitem{41NEURIPS2022_38e511a6}
\BIBentryALTinterwordspacing
J.~Liao, Y.~Ding, Y.~Shavit, D.~Huang, S.~Ren, J.~Guo, W.~Feng, and K.~Zhang, ``Wt-mvsnet: Window-based transformers for multi-view stereo,'' in \emph{Advances in Neural Information Processing Systems}, S.~Koyejo, S.~Mohamed, A.~Agarwal, D.~Belgrave, K.~Cho, and A.~Oh, Eds., vol.~35.\hskip 1em plus 0.5em minus 0.4em\relax Curran Associates, Inc., 2022, pp. 8564--8576. [Online]. Available: \url{https://proceedings.neurips.cc/paper_files/paper/2022/file/38e511a690709603d4cc3a1c52b4a9fd-Paper-Conference.pdf}
\BIBentrySTDinterwordspacing

\bibitem{42NEURIPS2022_94ef7217}
\BIBentryALTinterwordspacing
J.~Zhang, R.~Tang, Z.~Cao, J.~Xiao, R.~Huang, and L.~FANG, ``Elasticmvs: Learning elastic part representation for self-supervised multi-view stereopsis,'' in \emph{Advances in Neural Information Processing Systems}, S.~Koyejo, S.~Mohamed, A.~Agarwal, D.~Belgrave, K.~Cho, and A.~Oh, Eds., vol.~35.\hskip 1em plus 0.5em minus 0.4em\relax Curran Associates, Inc., 2022, pp. 23\,510--23\,523. [Online]. Available: \url{https://proceedings.neurips.cc/paper_files/paper/2022/file/94ef721705ea95d6981632be62bb66e2-Paper-Conference.pdf}
\BIBentrySTDinterwordspacing

\bibitem{16zhang2023geomvsnet}
Z.~Zhang, R.~Peng, Y.~Hu, and R.~Wang, ``Geomvsnet: Learning multi-view stereo with geometry perception,'' in \emph{Proceedings of the IEEE/CVF Conference on Computer Vision and Pattern Recognition}, 2023, pp. 21\,508--21\,518.

\bibitem{35qi2017pointnet}
C.~R. Qi, H.~Su, K.~Mo, and L.~J. Guibas, ``Pointnet: Deep learning on point sets for 3d classification and segmentation,'' in \emph{Proceedings of the IEEE conference on computer vision and pattern recognition}, 2017, pp. 652--660.

\bibitem{17kendall2017end}
A.~Kendall, H.~Martirosyan, S.~Dasgupta, P.~Henry, R.~Kennedy, A.~Bachrach, and A.~Bry, ``End-to-end learning of geometry and context for deep stereo regression,'' in \emph{Proceedings of the IEEE international conference on computer vision}, 2017, pp. 66--75.

\bibitem{21he2015spatial}
K.~He, X.~Zhang, S.~Ren, and J.~Sun, ``Spatial pyramid pooling in deep convolutional networks for visual recognition,'' \emph{IEEE transactions on pattern analysis and machine intelligence}, vol.~37, no.~9, pp. 1904--1916, 2015.

\bibitem{20kingma2014adam}
D.~P. Kingma and J.~Ba, ``Adam: A method for stochastic optimization,'' \emph{arXiv preprint arXiv:1412.6980}, 2014.

\end{thebibliography}

\end{document}